\documentclass[journal]{IEEEtran}
\usepackage{cite}
\usepackage{times,amsmath,amssymb,psfrag,stfloats}
\usepackage{tabularx}
\usepackage{graphicx}
\usepackage{epsfig}
\usepackage{epstopdf}
\graphicspath{{./pics/}} \DeclareGraphicsExtensions{.pdf,.jpg,.png,.eps}
\usepackage{color}
\usepackage{algorithmic}
\usepackage{algorithm}
\usepackage{hyperref}
\usepackage{multirow}
\usepackage{caption}
\usepackage{flushend}

\renewcommand{\baselinestretch}{1}

\begin{document}
\title{Enhancement of Power Equipment Management Using Knowledge Graph}

\author{\IEEEauthorblockN{Yachen Tang\IEEEauthorrefmark{1},
Tingting Liu\IEEEauthorrefmark{1}, Guangyi Liu\IEEEauthorrefmark{1}, Jie Li\IEEEauthorrefmark{2}, Renchang Dai\IEEEauthorrefmark{1}, and Chen Yuan\IEEEauthorrefmark{1}}\\
\IEEEauthorblockA{\IEEEauthorrefmark{1}GEIRI North America, San Jose, CA, USA}\\
\IEEEauthorblockA{\IEEEauthorrefmark{2}State Grid US Representative Office, New York City, NY, US.}

\thanks{This work is supported by the State Grid Corporation technology project 5455HJ180022.}}

\maketitle

\begin{abstract}
Accurate retrieval of the power equipment information plays an important role in guiding the full-lifecycle management of power system assets. Because of data duplication, database decentralization, weak data relations, and sluggish data updates, the power asset management system eager to adopt a new strategy to avoid the information losses, bias, and improve the data storage efficiency and extraction process. Knowledge graph has been widely developed in large part owing to its schema-less nature. It enables the knowledge graph to grow seamlessly and allows new relations addition and entities insertion when needed. This study proposes an approach for constructing power equipment knowledge graph by merging existing multi-source heterogeneous power equipment related data. A graph-search method to illustrate exhaustive results to the desired information based on the constructed knowledge graph is proposed. A case of a 500 kV station example is then demonstrated to show relevant search results and to explain that the knowledge graph can improve the efficiency of power equipment management.
\end{abstract}

\begin{IEEEkeywords}
Power equipment management, knowledge graph, data merge, entity extraction, entities relationship.
\end{IEEEkeywords}
\IEEEpeerreviewmaketitle

\section{Introduction}
\IEEEPARstart{W}{ITH} the rapid development of power grid and the in-depth application of new technologies, information construction has placed higher demands on the platform capabilities of infrastructure. At the same time, power equipment manufacturing and management companies have a large number of suppliers, products, users and other information, and they are stored in different information systems, such as energy management system \cite{textbook}, asset management system \cite{8086023}, market management system, business operation support system, the auxiliary management system \cite{7917272, 6938947}, the quality management system \cite{chen1}, the security risk pre-control system \cite{6057498, 7731849}, and the financial management system, etc. In order to meet the needs of electrical business and technology development, an integrated enterprise-level information integration platform should be proposed to realize the smooth flow of data and information sharing between users and enterprises. The knowledge graph provides an effective solution to achieve this goal, especially in the field of power equipment management, the knowledge graph can improve planning, traceability \cite{6620157}, production scheduling, order execution, and supplier management of power equipment.

The existing knowledge graph technique is still in the initial stage. The concept of the knowledge graph was first proposed by Google in 2012 \cite{steiner_iswc_2012}. Its original intention was to optimize the performance of search engines. A knowledge graph is essentially an information management system that integrates all kinds of information and uses graphs to describe data and knowledge. The DBpedia \cite{bpdiamost} uses a fixed model to extract the entity information from Wikipedia, which including the abstract, infobox, category and page link. Yago \cite{Suchanek1242667} is a large-scale ontology that integrates Wikipedia and WordNet \cite{worldnet}. It first develops fixed rules to extract the infobox of each entity from Wikipedia, and then perform the ``Type Inference'' of each entity based on the category gathered from Wikipedia to obtain the ``Is a'' relationship between entities and concepts. BabelNet \cite{babel} is the largest multi-language synonym dictionary, which itself can be regarded as a ``Semantic Network'' composed of concepts, entities, and relationships. In the power industry, even though Siemens began to develop enterprise knowledge graph in 2017 \cite{SIEMENS} to solve problems such as data islands, data access difficulties, inefficient operation procedures \cite{8088673}, and low data quality, the knowledge graph has almost no application in power grid operation, monitoring, and equipment management.  

\begin{figure*}[h!bt]
\centerline{\includegraphics[scale = 0.5] {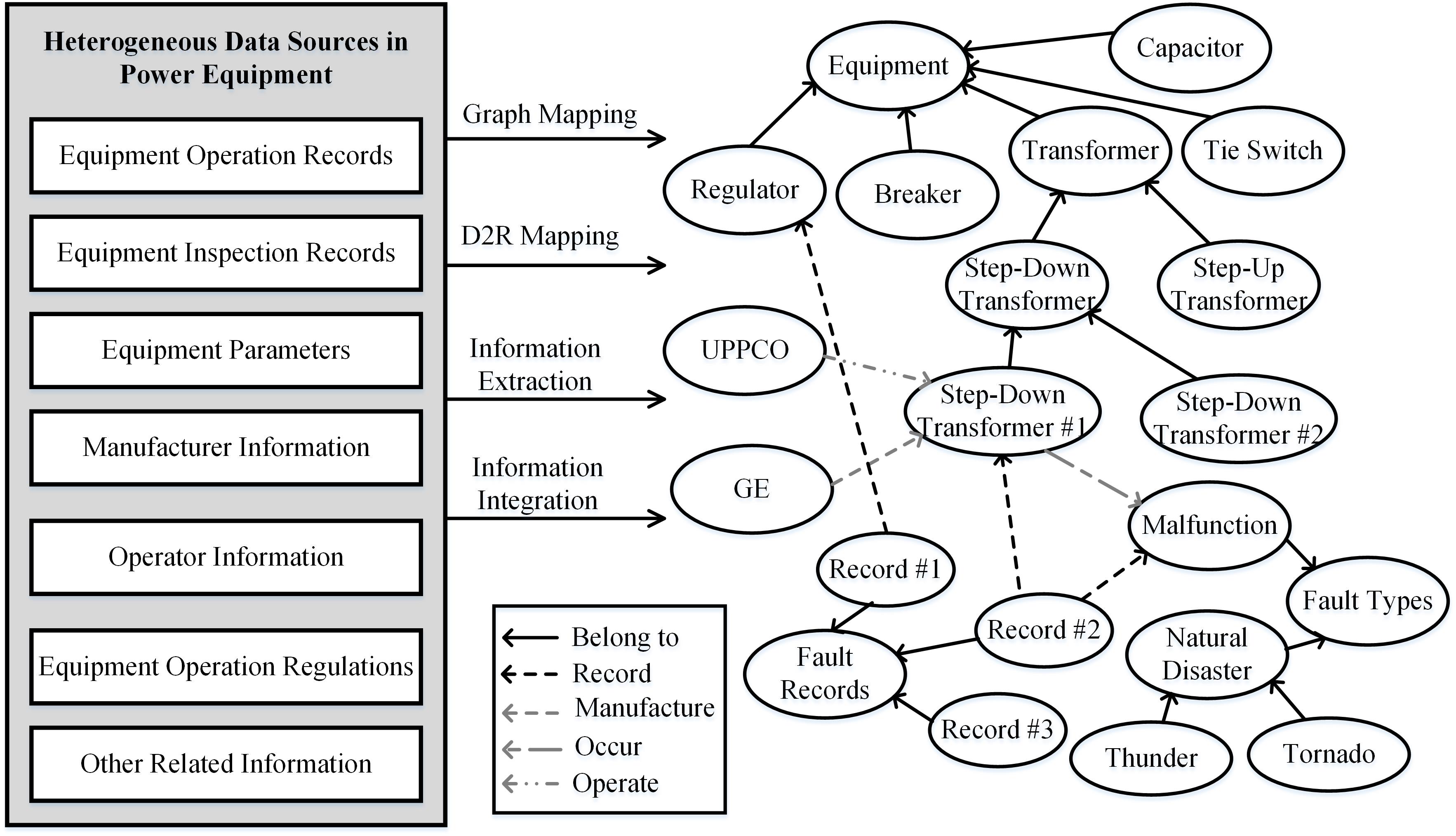}}
\caption{Schematic diagram of the establishment of an example power equipment knowledge graph.}\label{figure1}
\end{figure*}

The basic element of the knowledge graph is represented as a triple facts indicating the relation between two entities, i.e. entity 1, relation, entity 2, and it stores into the resource description framework (RDF) database \cite{kg1}. The data visualization of a complete knowledge graph is similar to a search tree structure, which is consist of a large number of basic elements as shown in Fig. \ref{figure1} depicting a schematic diagram of the establishment of an example power equipment knowledge graph. The knowledge graph is a graph-based data structure composed of nodes and edges. The node in this example represents an entity in power asset management system, including a power device, a fault type, a record sample, an operator, a manufacturer, etc., while the edge represents a relationship between one entity to another one, such as belonging to, record, causing, manufacture, occur, operate, and the like. The heterogeneous data sources for an integrated power equipment knowledge graph should include the historical equipment operation and inspection records, the detailed technical parameters of equipment, the comprehensive information of the relevant responsible person and/or department, equipment related specifications and standards, and all other correlative information. The knowledge entities and relations can be then extracted from the collected information via graph mapping \cite{kg2}, D2R mapping \cite{kg3}, and other data extraction and integration approaches. 

Power equipment manufacturing enterprises (industry groups) have a large number of information from suppliers, products, and users, while the electrical companies have abundant data from equipment operation. Because of the limitation of existing asset management system, the data exchange between equipment manufacturing enterprises and power companies are insufficient, which resulting in low efficiency of data usage. The power equipment knowledge graph proposed in this paper can construct a knowledge network of all power equipment in the specified power network topology and can query the basic information, classify relevant information of products, maintain the integrity of information, demonstrate the relationship between products, and can achieve real-time updates on the content of any product. This paper is organized as follows. Section II introduces the construction of knowledge graph for power equipment. Section III introduces the information searching process based on the power equipment knowledge graph. Section IV is a case study that demonstrates the search results from a station based on the proposed method. Section V concludes with future work.

\section{Knowledge Graph Creation for Power Equipment}
Knowledge graph is usually divided into two categories: general knowledge graph and enterprise knowledge graph. General knowledge is oriented to the whole field, mainly for Internet search, result recommendation, intelligent answers to questions, and other application scenarios, which is emphasizing the ``breadth'' of the graph database. The enterprise knowledge graph \cite{unknown} is domain-specific, and entities are professional terms at all levels in a particular industry that will provide high searching accuracy to assist with complex analytical applications or decision support. The power equipment knowledge graph belongs to a kind of enterprise knowledge graph.

\begin{figure}[h!bt]
\centerline{\includegraphics[scale = 0.5] {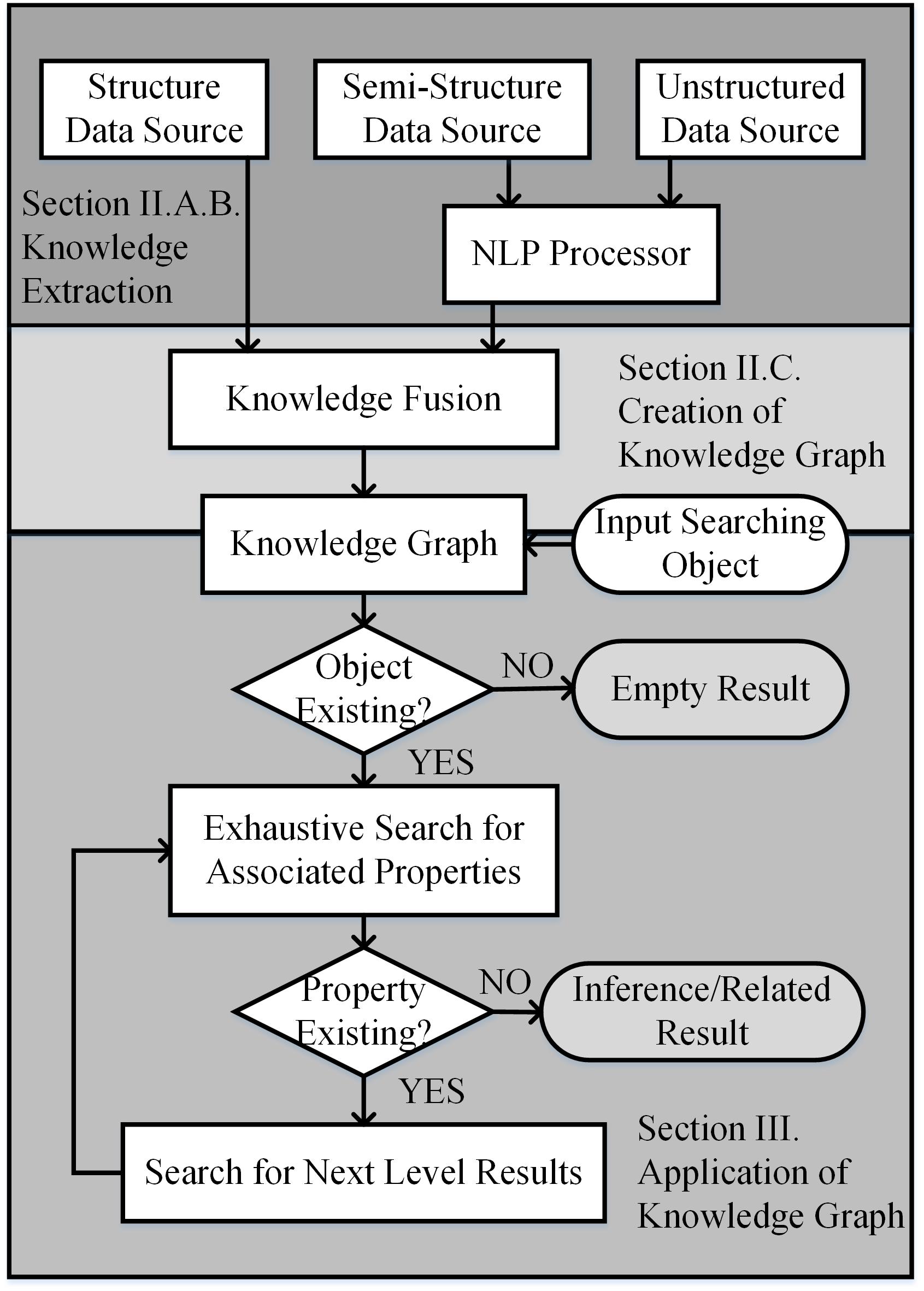}}
\caption{Organizational flowchart of the proposed knowledge graph application.}\label{flowchart}
\end{figure}

Fig. \ref{flowchart} shows the flowchart to present the framework for this proposed work. This study can be divided into two parts: the knowledge graph construction and its application. The process of knowledge graph creation usually consists of three steps: i) knowledge extraction, ii) knowledge fusion, and iii) knowledge processing. The main function of knowledge extraction is to extract the entities, attributes, and relationships contained in the semi-structured and/or unstructured data sources \cite{Chen2009} through the natural language processing (NLP) processor \cite{citeulike}. Knowledge fusion first performs entity disambiguation and coreference resolution on the extracted entities, then integrates entities, attributes, and relationships with the existing structured data to form the initial knowledge graph. Knowledge processing is a dynamic process. In the continuous application process of the knowledge graph, the knowledge graph is updated and revised in combination with the development and enrichment of the knowledge-base.

\subsection{Entity Extraction of Power Equipment}
The main task of entity extraction is to extract words representing entities and/or attributes in power equipment related information and perform part-of-speech tagging. The proposed study directly utilizes an electric power dictionary to match and extract entities form data sources. The first step of entity extraction is to segment all equipment relevant information into words based on a common words dictionary and the hidden Markov model (HMM) \cite{KROGH2001567}. During this process, the electric power dictionary is imported to assist to improve the segmentation accuracy. Then, the algorithm will match the divided words with the common words dictionary and the electric power dictionary, if the matching item can be retrieved, the word is extracted as the entity of the knowledge graph.

In this study, only the connection status, the operation, and inspection information of each power equipment are considered. All words can be tagged into 8 categories: 1) the noun describing the power equipment and/or component is denoted as ``E1''; 2) the noun describing the electrical companies, system operators, and/or system management organizations are expressed as ``E2''; 3) the noun describing the manufacturer is denoted as ``E3''; 4) the verb describing the connection status between each equipment is denoted as ``R1''; 5) the verb describing the actions occurs during the operational inspection process is denoted as ``R2''; 6) the verb describing the actions occurs during the manufacturing process is expressed as ``R3''; 7) the nouns describing the phenomena that occur during the operation and the inspection, such as natural disasters, switch fusing, equipment outages, etc., indicating as ``P''; 8) unmatched words will be neglected.

\subsection{Relation Extraction}
The main task of relation extraction is to identify whether there is a relationship between each entity and the corresponding relationship type, such as the affiliation, connection, and causality, etc. Define the relation types between entities as shown in Table \ref{relext}, the relation extraction process is converted into a classification problem. In this study, we only consider the relationship between electrical equipment and other entities. Before the relation identification, the entity-pairs to be classified are formed first and then match the entity-pairs with the word-specific combinations demonstrated in the table. The relationship between each word-pairs can be identified.

\begin{table}[h!bt]
\caption{Relation types of entities.}
\label{relext}
\centering
\begin{tabular}{|c|c|c|} 
\hline
\textbf{Entity A} & \textbf{Entity B} & \textbf{Possible Relation} \\ 
\hline
E1 & E1 & A (B) R1 B (A) \textbf{or }No Relation  \\ 
\hline
E1 & E2 & B R2 A~\textbf{or~}No Relation          \\ 
\hline
E1 & E3 & B R3 A~\textbf{or~}No Relation          \\ 
\hline
E1 & P  & A occurs B~\textbf{or~}No Relation      \\
\hline
\end{tabular}
\end{table}

\subsection{Knowledge Graph Creation for Power Equipment}
The equipment parameters, operation records, manufacture related information and the other information of electric power equipment usually represents in the form of a single sentence in natural language. In this study, the power equipment knowledge graph construction process should notice:
\begin{enumerate}
\item Since the entity is limited to the electrical field and the power industry has a clear terminology standard, the entity ambiguity problem is basically non-existent, neglecting the disambiguation step;
\item The amount of entities in a specific domain is relatively small. The coreference resolution should be performed before the relationship extraction that the same entity can obtain more relational training samples;
\item After the relation extraction, the relationship needs to be filtered to avoid relation redundancy, which may cause duplicate storage and affect the search result in the subsequent application;
\item The information integration step organizes and merges the extracted triples together with the triples contained in the structured data to form a power equipment management knowledge graph.
\end{enumerate}

\section{Power Equipment Information Retrieval}
The second part of the flowchart in Fig. \ref{flowchart} illustrates the information retrieval process in the application of the equipment knowledge graph. When the searching object is input to the created knowledge graph, a criteria is performed to judge if the object exists in the graph database. Once the object is detected, the exhaustive search for all associated properties (all related entities) to the object is performed. According to searching results, determine if there is a desired outcome and perform the inference process or the next level search.

\begin{figure*}[h!bt]
\centerline{\includegraphics[scale = 0.68] {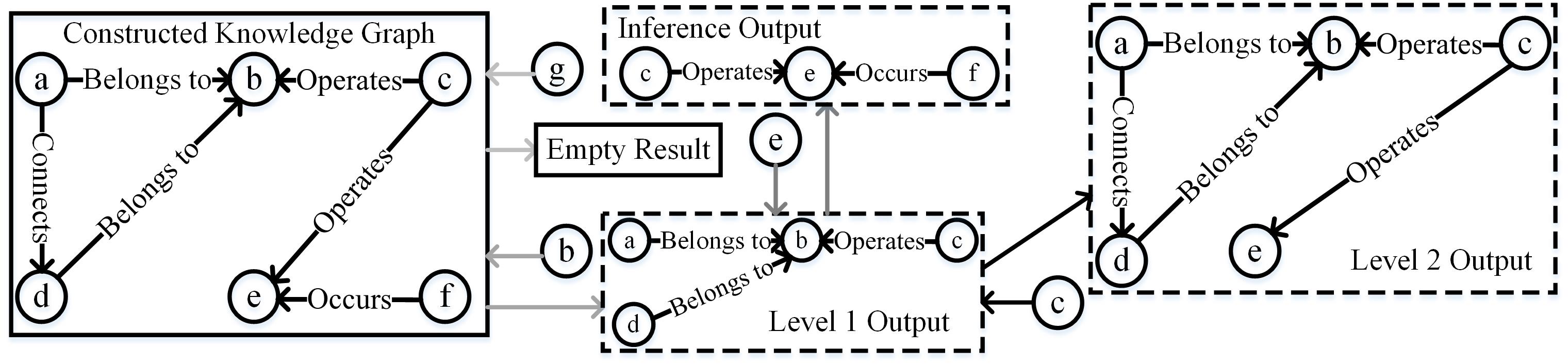}}
\caption{Example of information retrieval from the equipment knowledge graph.}\label{example}
\end{figure*}

Fig. \ref{example} shows an example of the information retrieval process from the constructed equipment knowledge graph. In this example, entities ``a'', ``b'', ``d'', and ``e'' belong to ``E1'', ``c'' is an element of ``E2'', and ``f'' belongs to ``P''. The relations ``Belongs to'' and ``Connects'' pertain to the type ``R1'' while ``Operates'' belongs to ``R2''. When the searching object ``g'' is input to the constructed knowledge graph, no searching result demonstrated since there is no this entity in the graph database. Making ``b'' as another input to the knowledge graph, the level 1 output illustrates all ``b-related'' entities and corresponding relations. When ``e'' is input to the level 1 result, the inference output shows ``e-related'' results and when ``c'' is input to the level 1 output, the next level result is shown in the rightmost in this figure.

The purpose of power equipment information retrieval is to find all relevant information in the created equipment knowledge graph that is consistent with the input search target. In a specific implementation, the given search target and the entity in the equipment knowledge graph can be matched one by one, and all entities and/or attributes related to the entity are output. From the perspective of the knowledge graph, if the input objective entity matches the storage entity in the graph database, it is equivalent to the corresponding entity path in the knowledge graph and consistent to the complete tree connected by the attribute node. Therefore, in the management of power equipment, when searching for a specific equipment or faulted-equipment \cite{7294716} that users want to check, the complete tree corresponding to the equipment's entity is needed to find only, and then perform deep search and traceability analysis on the complete tree to quickly locate the equipment, find out the possible problems, and contact the relevant responsible unit.

\section{Case Study in Power Equipment Management}
\begin{figure*}[h!bt]
\centerline{\includegraphics[scale = 0.48] {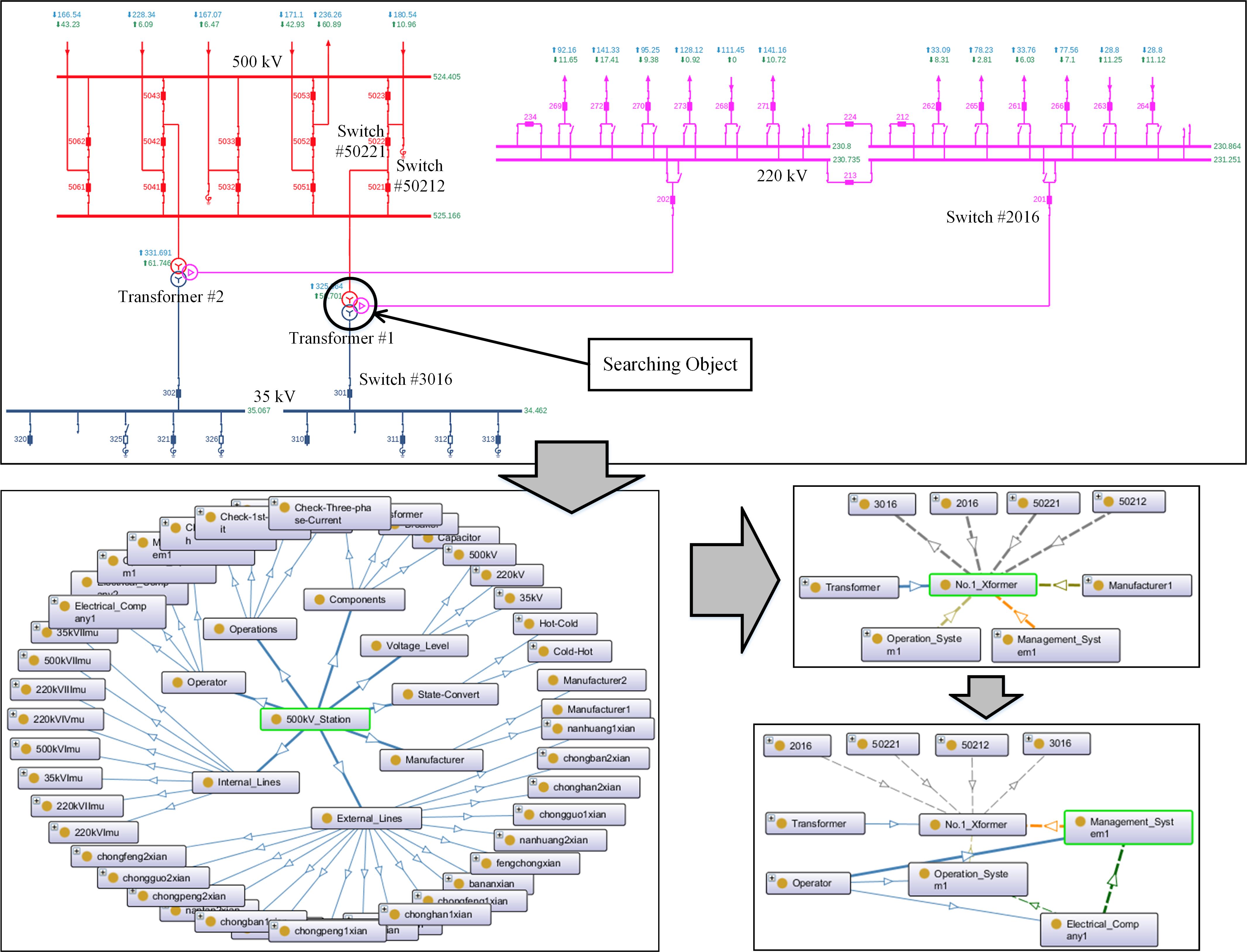}}
\caption{Case study of the power equipment knowledge graph from a 500kV station.}\label{case}
\end{figure*}

This section is to validate the proposed method with the realistic multi-sources data from a station. The single-line diagram of the tested station is shown in Fig. \ref{case}. The electrical components in this station mainly include two transformers, eight capacitors, 40 breakers, and 84 switches. The topological connection between each component can also be gathered from the diagram. The constructed RDF database merges the information of operators and manufacturers. In this study, we adopted the Protege \cite{protege}, which is an open-source ontology editor to achieve the visualization of the ontology graph. The knowledge graph representation is shown in the lower left part of Fig. \ref{case}, eight first-level class, ``Operator'', ``Operations'', ``Components'', ``Voltage-Level'', ``State-Convert'', ``Manufacturer'', ``External-Lines'', and ``Internal-Lines'', directly belong to the entity of ``500 kV Station''. Because of the security considerations, some entities in this case can only be represented by numbers.

Assume the searching object in this study is the ``Transformer \#1'', the level 1 output demonstrates all switches, ``\#2016'', ``\#3016'', ``\#50212'', and ``\#50221'', connected with this transformer, which can help the operator and manager to localize the equipment promptly. The relations between these four switches and this transformer are ``Connect'', which is shown with gray edges in the result. Also, the ``Transformer \#1'' is a subclass of ``Transformer'', and the blue edge in the output represents the ``Belong to'' relation. In addition, this transformer is operated (represented as the light yellow edge) by the ``Operation System 1'' and managed (represented as the orange edge) by the ``Management System 1''. Furthermore, this equipment is manufactured by ``Manufacturer 1'' and the relation between these two entities is represented by the dark green edge.

In the next level search, the management system is checked. The corresponding search result is shown in the lower right part of Fig. \ref{case}. This management system is controlled by the ``Electrical Company 1'' and the ``Control'' relation is represented by the green edge. The inference function in the knowledge graph also demonstrates the ``Control'' relation between the ``Electrical Company 1'' and the ``Operation System 1''. According to the output, the connection status, the management and controlled system, and the responsible units' information of the target transformer can be obtained promptly and accurately.

\section{Conclusion and Future Work}
This work represents the knowledge graph approach to enhance the management of power equipment in the electrical system that is conducted based on heterogeneous multi-sources datasets. The proposed knowledge graph method can provide an effective enhancement for the power equipment management, which includes the equipment operation, maintenance, fault process, replacement, procurement, traceability, etc. Words classification and recognition is the key to extracting entities and relationships from semi-structured and/or unstructured information. The graph database in RDF format is adopted to store the extracted data and utilized to construct a knowledge graph. The exhaustive search for all associated properties (all related entities) to the object based on the constructed database. A case study utilized the real data gathered from a 500 kV station to demonstrate the search result based on the knowledge graph. In the continuous application process of the knowledge graph, the graph database keeps updating and revising in combination with the development and enrichment of the knowledge-base. 

As a future work, we intend to gather more various data sources, such as social network information, operating guidelines and regulations for different power systems and equipment, environmental conditions, etc. to build a more comprehensive and dynamic graph database. According to the constructed knowledge graph, build a manufacturer evaluation mechanism to improve the quality of equipment production, propose equipment customization requirements in a special operating environment, introduce the equipment recommendation system, and try to establish a dedicated equipment trading platform for the power system. 

\bibliographystyle{IEEEtran}
\bibliography{IEEEabrv,RefDatabase}

\end{document}